\documentclass[conference]{IEEEtran}
\IEEEoverridecommandlockouts
\usepackage{amsmath,amsfonts}
\usepackage{array}
\usepackage{subcaption}
\usepackage{textcomp}
\usepackage{stfloats}
\usepackage{url}
\usepackage{makecell}
\usepackage{multirow}

\usepackage{verbatim}
\usepackage{graphicx}
\def\BibTeX{{\rm B\kern-.05em{\sc i\kern-.025em b}\kern-.08em
    T\kern-.1667em\lower.7ex\hbox{E}\kern-.125emX}}
\usepackage{balance}
\usepackage{ifpdf}
\usepackage{booktabs}

\usepackage{cite}
\usepackage{color}
\usepackage{verbatim}

\usepackage{makecell}
\usepackage{amsmath}
\usepackage{amsthm}
\usepackage{bbm}
\usepackage{bm}
\usepackage{amssymb}
\usepackage{algorithm}
\usepackage[noend]{algpseudocode}

\usepackage{array}

\usepackage{pifont}
\usepackage{color}

\usepackage{fixltx2e}

\ifCLASSOPTIONcaptionsoff
\usepackage[nomarkers]{endfloat}
\let\MYoriglatexcaption\caption
\renewcommand{\caption}[2][\relax]{\MYoriglatexcaption[#2]{#2}}
\fi

\usepackage{caption}
\begin{document}
\title{Edge Intelligence Optimization for Large Language Model Inference with Batching and Quantization}

\author{\IEEEauthorblockN{Xinyuan Zhang\IEEEauthorrefmark{1}, Jiang Liu\IEEEauthorrefmark{1}\IEEEauthorrefmark{2}, Zehui Xiong\IEEEauthorrefmark{3}, Yudong Huang\IEEEauthorrefmark{1}, Gaochang Xie\IEEEauthorrefmark{1}, Ran Zhang\IEEEauthorrefmark{1}\IEEEauthorrefmark{2}
		\thanks{This work was supported by the National Natural Science Foundation of China (Grant No. 62171064).}\thanks{The corresponding author is Jiang Liu (Email: liujiang@bupt.edu.cn).}	
	}
	\IEEEauthorblockA{\IEEEauthorrefmark{1}State Key Laboratory of Networking and Switching Technology, BUPT, China}
	\IEEEauthorblockA{\IEEEauthorrefmark{2}Purple Mountain Laboratories, Nanjing, China}
	\IEEEauthorblockA{\IEEEauthorrefmark{3}Information Systems Technology and Design Pillar, SUTD, Singapore}
}

\maketitle

\begin{abstract}

Generative Artificial Intelligence (GAI) is taking the world by storm with its unparalleled content creation ability. Large Language Models (LLMs) are at the forefront of this movement. However, the significant resource demands of LLMs often require cloud hosting, which raises issues regarding privacy, latency, and usage limitations. Although edge intelligence has long been utilized to solve these challenges by enabling real-time AI computation on ubiquitous edge resources close to data sources, most research has focused on traditional AI models and has left a gap in addressing the unique characteristics of LLM inference, such as considerable model size, auto-regressive processes, and self-attention mechanisms. In this paper, we present an edge intelligence optimization problem tailored for LLM inference. Specifically, with the deployment of the batching technique and model quantization on resource-limited edge devices, we formulate an inference model for transformer decoder-based LLMs. Furthermore, our approach aims to maximize the inference throughput via batch scheduling and joint allocation of communication and computation resources, while also considering edge resource constraints and varying user requirements of latency and accuracy. To address this NP-hard problem, we develop an optimal \textit{D}epth-\textit{F}irst \textit{T}ree-\textit{S}earching algorithm with online tree-\textit{P}runing (DFTSP) that operates within a feasible time complexity. Simulation results indicate that DFTSP surpasses other batching benchmarks in throughput across diverse user settings and quantization techniques, and it reduces time complexity by over 45\% compared to the brute-force searching method.

\end{abstract}

\begin{IEEEkeywords}
	
Generative AI, large language model, edge intelligence, wireless networks
\end{IEEEkeywords}

\section{Introduction}

We are living in an era of rapid Artificial Intelligence (AI) advancements. In the past year, Generative AI (GAI) has revolutionized the AI field, automating content creation and liberating creators from time-consuming manual efforts \cite{hacker2023regulating}. A standout within GAI is the Large Language Models (LLMs), such as GPT and Claude. Their applications extend to code generation, customer service chatbots, novel writing, and beyond \cite{zhao2023survey}. At its core, such impressive generative potential of LLMs arises from their complex architecture with billions of neurons, which leads to resource-intensive training and inference. Hence, LLMs are primarily hosted in the cloud. However, this centralized approach brings forth challenges like privacy concerns, usage limitations, and latency \cite{xu2023unleashing}, impeding GAI's broader acceptance and underscoring the need for innovative solutions.

Edge intelligence presents a viable solution to these challenges faced by LLMs, building on its proven success in traditional AI services. By leveraging ubiquitous edge computing resources near data sources, edge intelligence mitigates issues of prolonged propagation latency and resource limitations. With user data distributed across various edge nodes, rather than being solely stored on cloud servers, the risk of data leakage diminishes. The authors in \cite{8736011} delineated the advantages of edge intelligence, while also highlighting the challenges of executing efficient inference on resource-constrained edge nodes. The authors in \cite{9606540} proposed partitioning deep neural network (DNN) inference between local devices and edge servers, leveraging multi-threading to enhance DNN inference throughput. In \cite{9769868}, the authors jointly optimized the early-exit selection, model partitioning, and computation resource allocation to minimize the device-edge collaborate inference latency. While these studies provide valuable insights into edge intelligence optimization, they mainly focus on traditional DNN models, overlooking the unique challenges posed by LLMs to edge inference.


Optimizing edge intelligence tailored for transformer-based LLMs is challenging due to LLM's distinct attributes compared to traditional DNN models. The reasons are threefold. (1) Transformer-based LLMs are significantly larger in size. For instance, GPT-4 has around 1.8 trillion parameters \cite{GPT4}, while AlexNet only has 61 million \cite{8876870}. This surge in size notably increases both processing time and memory consumption. (2) Transformer decoder-based LLMs generate outputs sequentially in an autoregressive manner \cite{NIPS2017_3f5ee243}. Given an input sequence, LLM traverses the entire layer stack to produce only one output token. This procedure is repeated for each output token, intensifying the computational demands. (3) LLMs employ self-attention modules, which means each output token is computed based on all preceding tokens. Consequently, the memory cache for all prior tokens must be maintained until the output sequence concludes, further increasing memory demands \cite{NIPS2017_3f5ee243}. In essence, the distinctive attributes of LLMs make their inference on resource-constrained edge nodes challenging, which is not tackled in existing research.

In this paper, we introduce a novel edge intelligence optimization problem tailored for transformer decoder-based LLM inference within wireless edge networks. We utilize model quantization to store model weights and activations in decreased bit precisions, reducing the memory footprint on edge servers \cite{yao2023comprehensive}. Furthermore, we employ batching to process requests from multiple users in parallel \cite{10038543}, thereby boosting the inference throughput. We also account for the varied user requirements of latency and text generation accuracy. For instance, while dialogue-based interactions in the Metaverse demand strict timelines, online medical prescriptions
prioritize text accuracy. We introduce the perplexity differential \cite{yao2023comprehensive} as a metric to strike a balance between the latency reduction and accuracy loss due to quantization. 

This paper's contributions can be summarized as follows:
\begin{itemize}
	\item We propose an edge intelligence optimization problem for LLM inference in wireless networks, aiming to maximize inference throughput via batching scheduling and joint allocation of communication and computation resources. Constraints include communication and memory resources on edge devices and user-specific latency and accuracy demands. The problem is a variant of multi-dimensional knapsack problem, which is NP-hard.
	\item We design an optimal \textit{D}epth-\textit{F}irst \textit{T}ree-\textit{S}earching algorithm with tree-\textit{P}runing (DFTSP) to address the problem in a polynomial time. The solution space is constructed as a search tree. To expedite the solution process, searching prioritizes branches formed by requests with higher latency tolerance, shorter output length, and lower communication bandwidth demands. Tree-pruning eliminates searching redundant tree branches, streamlining the algorithm's complexity.
	\item In the simulation, we demonstrate the superior throughput performance of DFTSP against other batching schemes across various user settings and quantization methods. Moreover, the advantage of DFTSP algorithm is confirmed compared to the brute-force searching approach.
\end{itemize}
\vspace{-2pt}

The rest of the paper is structured as follows: Section \uppercase\expandafter{\romannumeral2} presents the system model and problem formulation. The solution is proposed in Section \uppercase\expandafter{\romannumeral3}. Simulation results are given in Section \uppercase\expandafter{\romannumeral4}, followed by the conclusions in Section \uppercase\expandafter{\romannumeral5}.

\section{System model and problem formulation}

Consider a wireless edge intelligence network as shown in Fig. 1. An edge node (EN) serves multiple users. The EN combines a base station and a nearby edge server connected via wired links. The EN hosts an LLM using model quantization. Notice that while Fig. 1 focuses on one LLM, our approach is adaptable for multiple LLMs. The EN's total uplink and downlink bandwidths are $B^U$ and $B^D$, with $C$ and $M$ indicating EN's computing speed and memory capacity, respectively. User requests are indexed by $i \in \mathcal{I} = \left\{1,2, ..., I \right\}$. As shown in Fig. 1, the inference request information $\langle s_i, n_i, \tau _i, a_i \rangle$ is sent to EN via application APIs, just like ChatGPT playground \cite{chatGPTplayground}. Here, $s_i$ is $i$'s input prompt length, $n_i$ is $i$'s desired maximum text output length, categorized into multiple levels of $\left\{N_1, N_2, ..., N\right\}$, depending on the input length and the type of LLM service\footnote{ For example, in translation, the output length matches the input. For text summary, the output is much shorter than the input, while in article writing, the output is much longer.}. We assume the maximum output length is set by users or the API. $\tau _i$ is the latency requirement, and $a_i$ is the required text output accuracy. 

The following protocol for LLM inference on the EN is assumed. As shown in Fig. 2, time is divided into epochs, each of which is further divided into an uplink communication slot $T_U$, a computation slot $T_C$, and a downlink communication slot $T_D$. For resource efficiency, each $T_C$ starts immediately after the prior, overlapping with the preceding epoch's $T_D$ and the subsequent epoch's $T_U$. Slot durations are periodically updated based on long-term observation. EN aggregates requests that arrive in the previous epoch, and schedules LLM inference for the next epoch. Once scheduled, EN allocates resources for the assigned requests and users get notified. 


\begin{figure}[!t]
	\centering
	\includegraphics[width=2.5in]{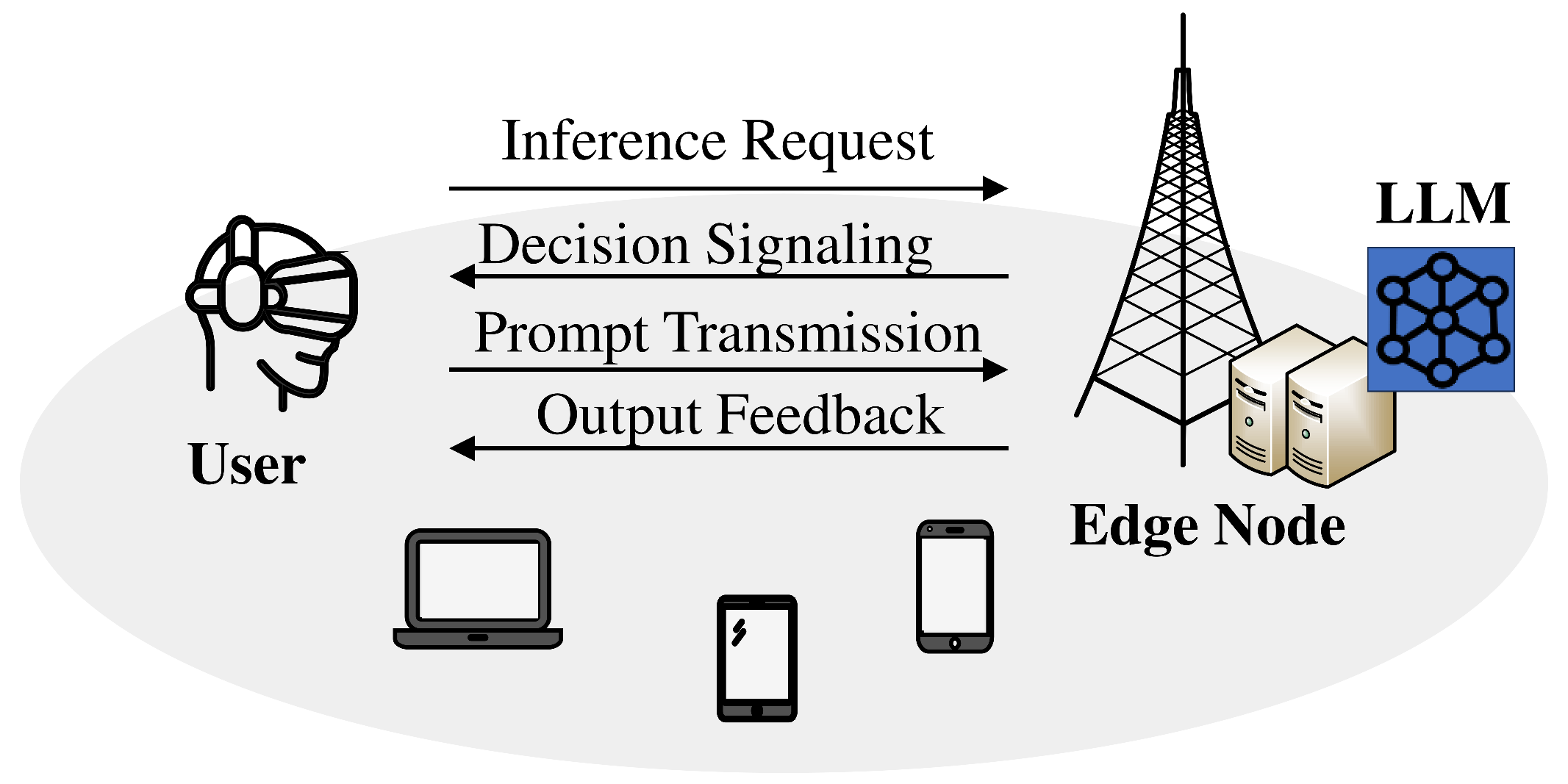}
	\vspace{-3pt}
	\caption{The wireless edge intelligence network and workflows for LLM inference. An LLM is deployed with model quantization.}
	\vspace{-8pt}
\end{figure}

\subsection{Communication Model}
\vspace{-3pt}

The data sizes for inference request and decision signaling in Fig. 1 are negligible. The communication model focuses on prompt transmission and output feedback. The system uses orthogonal frequency-division multiple access (OFDMA) for broadband spectrum allocation to users. Given the vast number of sub-carriers (e.g., thousands in 5G), bandwidth splitting can be viewed as continuous \cite{10038543}. Let $\rho _{i}^U$ and $\rho _{i}^D$ denote the bandwidth fractions for $i$'s allocated uplink and downlink, respectively, both between $0$ and $1$. For any scheduled user request $i$, the channel is frequency non-selective \cite{10038543}, with channel gain $h_{i}$ being constant within an epoch and can get known by the EN via measurement techniques like CSI-RS. The uplink and downlink transmission power are $p_{i}^U$ and $p^D$, respectively. The transmission rate are given by:
\vspace{-6pt}
\begin{equation*}
\vspace{-6pt}
r_{i}^U\! \!=\! \rho _{i}^U\! B^U\!\! \log_{2}\! {\left(\! 1\!+\! \frac{p_{i}^U {h_{i}}^2}{N_0}\! \right)}, r_{i}^D \!\!=\! \rho _{i}^D \!B^D\! \log_{2}\! {\left(\! 1\!+\! \frac{p^D {h_{i}}^2}{N_0} \!\right)}, 
\end{equation*}
where $N_0$ is the white Gaussian channel noise power. Considering the prompt must be uploaded within $T_U$, represented mathematically as $r_{i}^U T_U \ge s_i$, the minimum fraction of allocated uplink bandwidth $\rho _{i, min}^U$ is:
\vspace{-5pt}
\begin{equation*}
\vspace{-5pt}
\rho _{i, min}^U \triangleq \frac{s_i}{T_U B^U \log_{2}{\left( 1+ \frac{p_{i}^U {h_{i}}^2}{N_0}\right)}},
\end{equation*}
where $\!\rho_{i}^U\!\! \ge \!\rho _{i, min}^U\!$ if $i$ is scheduled. Similary, $\! \rho _{i, min}^D\!$ is defined.
\begin{figure}[!t]
	\centering
	\includegraphics[width=3.1in]{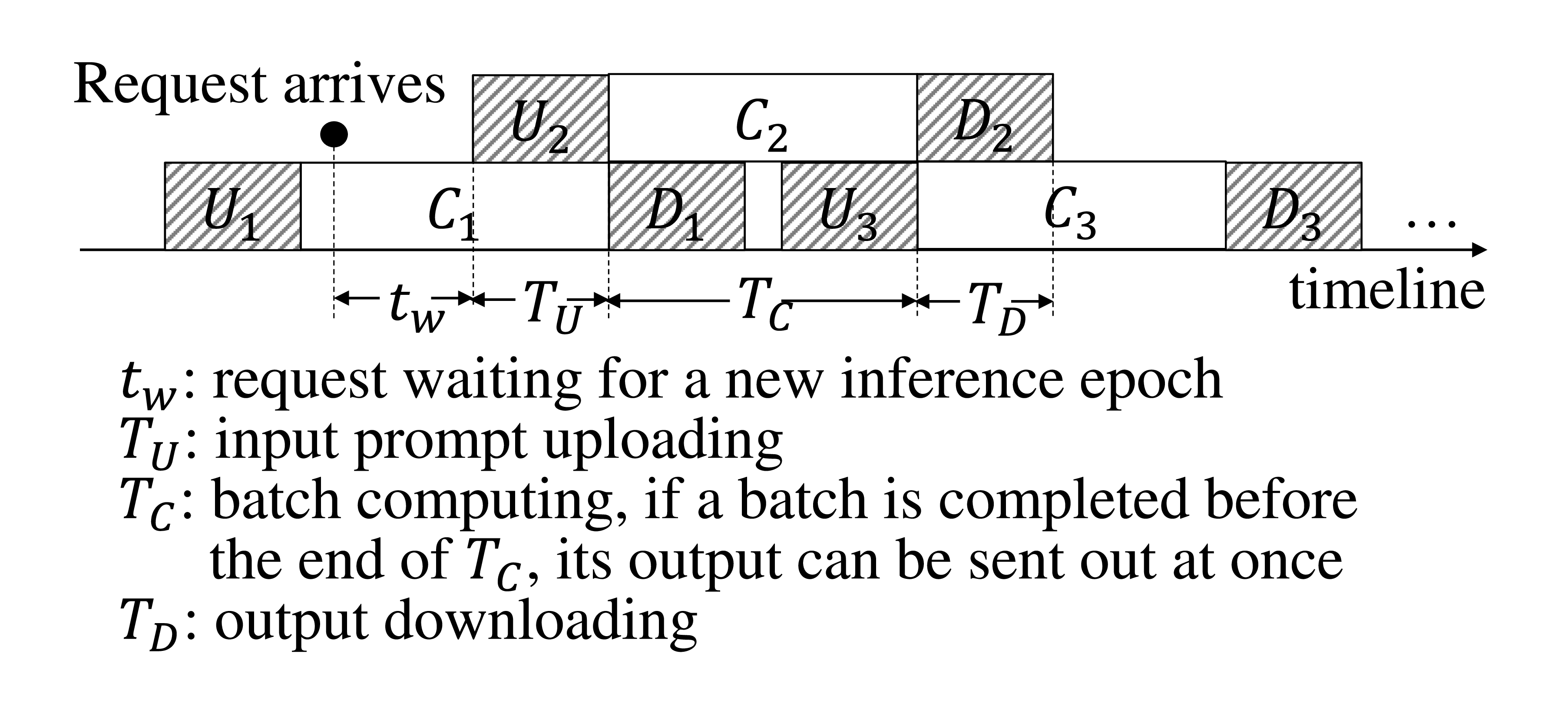}
	\vspace{-5pt}
	\caption{Timeline for LLM edge inference.}
	\vspace{-10pt}
\end{figure}

\vspace{-7pt}
\subsection{Inference Model}
\vspace{-3pt}

In this part, we analyze the memory footprint and latency of the batched LLM inference based on the transformer decoder-based computation precedures. 

LLMs, built with stacked transformer layers, operate in an autoregressive manner. Fig. 3 depicts a three-layer GPT-3's simplified inference. We use GPT-3 for example as it exemplifies transformer decoder-based LLMs, with GPT-4 details unavailable. As shown, a sequence of prompt tokens undergoes all layers to generate an output token, which is then fed back and processed again until an end-of-sequence token $\langle EOS \rangle$ is generated. Each full run through all layers is an iteration. The first iteration, termed the \textit{Initial Stage}, processes all prompt tokens to produce the first output token. Subsequent iterations, termed the \textit{Auto-regressive Stage}, use the previously generated tokens to produce subsequent tokens \cite{yu2022orca}. 

LLM inference's memory footprint mainly comes from weight storage and key-value cache (KV cache). The KV cache, essential for the \textit{Attention} mechanism, facilitates inter-token awareness by storing the matrices of token keys and values for weighted token averaging \cite{sheng2023high}. In the following, we formulate the memory footprint for weight storage and KV cache, as well as the latency during \textit{Initial Stage} and \textit{Auto-regressive Stage}.

\textit{(1) Memory footprint for weight storage:} Let $d_m$ represent the transformer's hidden dimension and $d_f$ the hidden dimension of Feed-Forward Network (FFN). Assuming each parameter is stored using a 2-byte floating point format, a matrix $\mathbf{A} \in \mathcal{R}^{ m \times n}$ requires $2mn$ bytes. Focusing on the $l$-th transformer decoder layer, the weight parameters are specified by $\mathbf{w}_Q^l,\! \mathbf{w}_K^l,\! \mathbf{w}_V^l \!\! \in\!\! \mathcal{R}^{d_m \!\times d_m}\!$, $\!\mathbf{w}_O^l\! \!\in\!\! \mathcal{R}^{ d_m \!\times d_m}\!$, $\!\mathbf{w}_1^l\!\! \in \!\!\mathcal{R}^{ d_m \!\times d_f}$, and $\mathbf{w}_2^l\! \in\! \mathcal{R}^{ d_f \!\times d_m} $. Hence, the memory footprint for weight storage on GPUs during LLM inference is:
\vspace{-5pt}
\begin{equation*}
m_1 = L \left( 8d_md_hn_h +4d_md_f \right),
\vspace{-5pt}
\end{equation*}
where $L$ is the number of model layers.

\textit{(2) Memory footprint and latency during Initial Stage:} Before this stage, all input prompts must be extended to the maximum token length for parallel execution. We set the maximum length as $s'$. Given one input prompt of $\mathbf{X}^l \in \mathcal{R}^{ s' \times d_m}$, the computation operations of the \textit{Initial Stage} in Fig. 3 are:
\vspace{-4pt}
\begin{equation*}
\vspace{-8pt}
\mathbf{X}_V^l\!=\!\mathbf{X}^l \!\cdot\! \mathbf{w}_V^l,\ \mathbf{X}_K^l\!=\!\mathbf{X}^l \!\cdot\! \mathbf{w}_K^l,\ \mathbf{X}_Q^l\!=\!\mathbf{X}^l \!\cdot\! \mathbf{w}_Q^l, 
\end{equation*}
\begin{equation*}
\vspace{-8pt}
\mathbf{X}_{Out}^l=f_{\text{softmax}}(\frac{\mathbf{X}_Q^l{\mathbf{X}_K^l}^T}{ \sqrt{d_h}})\! \cdot\! \mathbf{X}_V^l\!\cdot\! \mathbf{w}_O^l \!+\! \mathbf{X}^l, 
\end{equation*}
\begin{equation*}
\vspace{-4pt}
\mathbf{X}^{l+1}=f_{\text{relu}}(\mathbf{X}_{Out}^l \cdot \mathbf{w}_1^l ) \cdot \mathbf{w}_2^l +\mathbf{X}_{Out}^l.
\end{equation*}

Consider the batched inference within any $T_C$, we define a binary variable $x_{i}$ to represent whether user $i$'s request is scheduled: $x_{i}=1$ if scheduled, and $x_{i}=0$ otherwise. The batch size is then $\sum_{i \in \mathcal{I}}x_{i}$. During the \textit{Initial Stage}, the size of KV cache ($\mathbf{X}_K^l, \mathbf{X}_V^l$) for all input prompts across layers is:
\vspace{-9pt}
\begin{equation*}
m_2^I = 4Ls'd_m \sum\nolimits_{i \in \mathcal{I}} x_{i}.
\vspace{-4pt}
\end{equation*}
\begin{figure}[!t]
	\centering
	\includegraphics[width=3.1in]{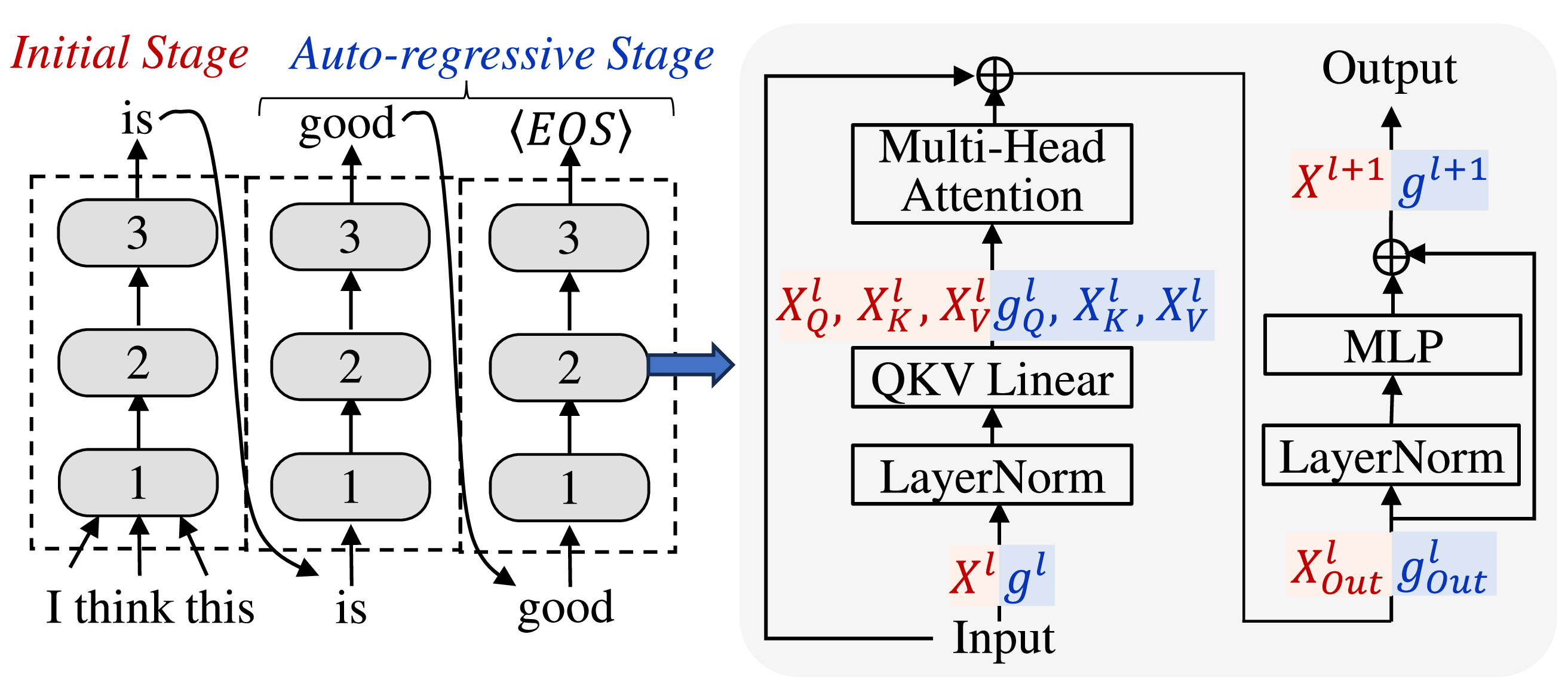}
	\vspace{-6pt}
	\caption{LLM inference procedure.}
	\vspace{-10pt}
\end{figure}
Floating point operations (FLOPs) quantify computational work. For a matrix-vector multiplication with $\mathbf{A} \in \mathcal{R}^{ m \times n},\mathbf{b} \in \mathcal{R}^{n}$, the computation requires $2mn$ FLOPs. The matrix-matrix multiplication involving $\mathbf{A} \in \mathcal{R}^{ m \times n},\mathbf{B} \in \mathcal{R}^{n \times p}$ needs $2mnp$ FLOPs. Hence, the batched inference latency for all input prompts across layers during the \textit{Initial Stage} is:
\vspace{-5pt}
\begin{equation*}
t^I \!\! =\! \frac{L\! \sum_{i \in \mathcal{I}}\! x_{i}}{C} \big(6s'\!d_m^2 \! + \!\big(4{s'}^2\!d_m \!+ 2s'\!d_m^2\big)\! +4s'\!d_md_f\!\big),
\vspace{-5pt}
\end{equation*}
where $6s'\!d_m^2$ denotes the computation to get $\mathbf{X}_Q^l,\! \mathbf{X}_K^l,\! \mathbf{X}_V^l$, $4{s'}^2\!d_m \!+ 2s'\!d_m^2$ denotes the calculation of $\mathbf{X}_{Out}^l$, and $4s'\!d_md_f$ is related to $\mathbf{X}^{l+1}$.

\textit{(3) Memory footprint and latency during Autoregressive Stage:} Given the input of $\mathbf{g}^l \!\!\!\in\!\! \!\mathcal{R}^{ 1 \times d_m}$, the computation operations in the $l$-th layer are:
\vspace{-3pt}
\begin{equation*}
\vspace{-8pt}
\mathbf{X}_V^l\!\gets\text{Concat}(\mathbf{X}_V^l, \mathbf{g}^l \!\cdot \!\mathbf{w}_V^l), \ \mathbf{X}_K^l\!\gets\text{Concat}(\mathbf{X}_K^l, \mathbf{g}^l\! \cdot\! \mathbf{w}_K^l), 
\end{equation*}
\vspace{-8pt}
\begin{equation*}
\vspace{-6pt}
\mathbf{g}_Q^l\!=\!\mathbf{g}^l\! \cdot\! \mathbf{w}_Q^l,\ \mathbf{g}_{Out}^l\!=\!f_{\text{softmax}}(\mathbf{g}_Q^l{\mathbf{X}_K^l}^T \!/\! \sqrt{d_h})\!\cdot\! \mathbf{X}_V^l\!\cdot\! \mathbf{w}_O^l\! + \mathbf{g}^l, 
\end{equation*}
\vspace{-8pt}
\begin{equation*}
\vspace{-3pt}
\mathbf{g}^{l+1}\!=\!f_{\text{relu}}(\mathbf{g}_{Out}^l\! \cdot\! \mathbf{w}_1^l )\! \cdot\! \mathbf{w}_2^l +\mathbf{g}_{Out}^l.
\end{equation*}
During this stage, the KV cache of $\mathbf{g}_K^l, \mathbf{g}_V^l$ updates with every new token until the sequence completes. Hence, the memory footprint of KV cache is:
\vspace{-7pt}
\begin{equation*}
\vspace{-5pt}
m_2^A = 4Ln_id_m \sum\nolimits_{i \in \mathcal{I}} x_{i}.
\end{equation*}
The batched inference latency of all output tokens across layers during \textit{Auto-regression Stage} is:
\vspace{-4pt}
\begin{equation*}
t^A = \frac{L}{C} \! \sum\nolimits_{i \in \mathcal{I}}\! x_{i} (n_i-1)\big(6d_m^2
+ \big(4(s'+\frac{n_i}{2})d_m+ 2d_m^2\big) +4d_md_f\big),
\vspace{-4pt}
\end{equation*}
where the latency increases with output iteration rounds.

\textit{3) Quantization Model}

Due to the resource demands of LLMs, the post-training quantization (PTQ) method is mainly adopted, as it streamlines computation without retraining. PTQ encodes LLM weights and activation tensors at precision below 16-bit, typically using 8-bit, 4-bit, or a combination \cite{zhao2023survey}. While some PTQ methods only adjust weights, others quantize both weights and activations. To quantify these effects, we define $\alpha$ to represent memory savings and $\beta$ to denote computational time reduction. Both metrics are measured via offline exhaustive evaluations on diverse datasets \cite{yao2023comprehensive}. 

Model quantization reduces memory usage and latency but can degrade accuracy. To address this, we use the \textit{perplexity} (PPL) differential, denoted as $\!\Delta$\textit{PPL}, to measure the change in LLM accuracy between pre- and post-quantization \cite{yao2023comprehensive}. Based on various datasets, $\!\Delta$\textit{PPL} is predetermined and known. A larger $\!\Delta$\textit{PPL} indicates a greater accuracy loss. We introduce $\!f\!$ as a function linking PPL differential to accuracy, which is monotonically decreasing. For users seeking high-accuracy LLM services, if $x_{i}\!=\!1$, it is imperative that $a_i\! \le \!f(\Delta PPL)$, ensuring generated text meets accuracy requirements.

\subsection{Problem Formulation}
\vspace{-2pt}

We formulate the edge intelligence optimization problem for LLM inference in wireless networks, aimed at maximizing throughput while adhering to resource allocation constraints and accommodating heterogeneous user QoS requirements. Let $\bm{X}\!=\!\left\{ x_{i} | \forall i \in \mathcal{I} \right\}$ be the decision variable, the optimization problem is formulated as follows:
\vspace{-5pt}
\begin{align}
\text{P1:} \ &  \max_{\bm{X}} \sum\nolimits_{i\in \mathcal{I}} x_{i}
\\
\text{s.t.} \quad & \sum\nolimits_{i\in \mathcal{I}} \rho_{i, min}^U x_{i}\le 1, \quad \sum\nolimits_{i\in \mathcal{I}} \rho_{i, min}^D x_{i}\le 1 \nonumber \tag{1a, 1b}
\\
& \alpha (m_1+m_2^I+m_2^A) \le M, \nonumber \tag{1c}
\\
& t_{w,i}+T_U+\beta (t^I+t^A) +T_D \le \tau_i, \forall x_{i} = 1, \nonumber \tag{1d}
\\
& a_i \le f(\Delta PPL), \quad \forall i \in \mathcal{I}, \nonumber \tag{1e}
\\
& x_{i} \in \left\{ 0,1 \right\}, \quad \forall i \in \mathcal{I}.  \nonumber \tag{1f}
\end{align}

\vspace{-5pt}
\noindent where the objective is the throughput of edge inference, measured by the number of user prompts processed successfully. (1a)-(1c) represent the constraints for uplink communication, downlink communication, and memory footprint, respectively. (1d) and (1e) ensure that each scheduled request receives satisfactory generation output within its deadline. By removing constraints (1d) and (1e), Problem P1 is simplified to a standard \textit{M}ulti-dimensional \textit{K}napsack \textit{P}roblem (MKP), which is NP-hard \cite{cacchiani2022knapsack}. The complexity of P1 is aggravated by introducing (1d) and (1e). Hence, P1 is proved to be NP-hard.

\section{Problem Solution}

In this section, we first reformulate the problem into a series of sub-problems aimed at finding feasible solutions. Then, we design an optimal DFTSP algorithm for resolution. We delve into the tree construction, searching, and pruning processes. Lastly, we analyze the algorithm's complexity.

\subsection{Problem Reformulation}

\begin{algorithm}
	\caption{ Optimal DFTSP algorithm.} 
	\begin{algorithmic}[1] \small
		\Require Available request set $\mathcal{\tilde{I}}$.
		\Ensure Optimal solution $\mathcal{S}$ to Problem P1.
		
		\State Initialize $z=0$, $d=0$, $\mathcal{F}_d = \emptyset $, $\mathcal{S}' = \emptyset $;
		\For{$z=| \tilde{\mathcal{I}} |, | \tilde{\mathcal{I}}  |-1, ..., 1$ }
		\State Sort $\mathcal{\tilde{I}}$ according to $\tilde{\tau_{i}}$'s descending order;
		\For{$d=z,z+1, ...,  | \tilde{\mathcal{I}}  |$}
		\State $\tau_{min}\gets \tilde{\tau_{d}}$, $\mathcal{F}_d \gets$ the first $d$ requests in $\mathcal{\tilde{I}}$;
		\State Construct root node $v_0$;
		\State Call DFS($v_0$, d);
		\If{DFS($v_0$, d) returns solution $\mathcal{S}'$}
		\State $\mathcal{S}  \gets \mathcal{S}'$, return $\mathcal{S}$;
		\EndIf
		\EndFor
		\EndFor	
		\Return no solution
		\Function{DFS}{$v_{N(v)},d$}
		\State $N(v) \gets$ depth of $v_{N(v)}$;
		\State Visit the path to $v_{N(v)}$ along unvisited nodes;
		\If{$\sum_{k=1}^{N(v)} v_k = d$}
		\State Recover the subset $\mathcal{S}'$ from $v_{N(v)}$;
		\If{$\mathcal{S}'$ meets constraints (2b)-(2e)}
		\State \Return $\mathcal{S}'$
		\Else
		\State Mark $v_{N(v)}$ visited;
		\State $v'\!\! \gets\!\! v_{N(v)}$'s sibling node with the largest index;
		\State Call DFS($v',d$);
		\EndIf
		\ElsIf{$\sum_{k=1}^{N(v)} v_k < d$ and $N(v)=N$}
		\State Mark $v_{N(v)}$, its parent node, and its sibling nodes 
		\Statex $\qquad \ $ with lower index visited;
		\State $v' \gets  v_{N(v)}$'s parent node;
		\State Call DFS($v',d$);
		\ElsIf{$\sum_{k=1}^{N(v)} v_k < d$ and $N(v)<N$}
		\If{all child nodes are visited} 
		\If{$v_{N(v)}$ is the root node}
		\State \Return no solution
		\Else
		\State $v' \gets  v_{N(v)}$'s parent node;
		\State Call DFS($v',d$);
		\EndIf
		\EndIf
		\State $v_{N(v)+1} \gets \min \left\{z', \left| \mathcal{F}_{N(v)+1} \right| \right\}$;
		\State Call DFS($v_{N(v)+1},d$);
		\EndIf
		\EndFunction
	\end{algorithmic}
\end{algorithm}

We observe that the optimization object is only concerned with the number of scheduled requests, whereas the constraints emphasize which requests are chosen. Let $\tilde{I}$ be the set of requests who are satisfied with the edge node's output accuracy, $\mathcal{S} \!=\! \left\{\!i |i\! \in\! \tilde{I}, x_i\! =\! 1\! \right\}$ denotes the set of scheduled requests, we have $\left | \mathcal{S} \right |\! =\! \sum_{i \in \tilde{\mathcal{I}}}x_i$. Problem P1 then translates to identifying such an $\mathcal{S}$ that possesses the maximum cardinality while satisfying the constraints. Given $\left | \mathcal{S} \right | = z$, Problem P1 is reformulated as the subsequent question:
\vspace{-5pt}
\begin{align}
\text{P2}: \quad & \text{find} \ \mathcal{S} \subseteq \mathcal{\tilde{I}} 
\\
\text{s.t.} \quad & \left | \mathcal{S} \right | = z, \nonumber \tag{2a}
\\
& \sum\nolimits_{i\in \mathcal{S}} k_is_i \le 1,\ \sum\nolimits_{i\in \mathcal{S}} k_1 n_i \le 1, \nonumber \tag{2b,2c}
\\
& \sum\nolimits_{i\in \mathcal{S}} n_i\le \tilde{M}, \nonumber \tag{2d}
\\
& \sum\nolimits_{i\in \mathcal{S}} k_4n_i + k_5 n_i^2 \le \tilde{\tau_{i}} \nonumber \tag{2e}.
\end{align}

\noindent Here, we suppose that the users are geographically concentrated, such as in the same building. In this case, $h_i$ is same for all users. Besides, $T_U$, $B^U$, $T_D$, and $B^D$ are known aforehand. Hence, $k_i$ is a variable associated with $T_U$, $B^U$, $p_i^U$, and $h_i$, and it varies with $p_i^U$. $k_1$ is a constant related to $T_D$, $B^D$, $p^D$, and $h_i$. $\tilde{M} \!\!= \!\!k_2 \!-\! s'z$, $\tilde{\tau_{i}}\! =\! \frac{(\tau _i-t_{w,i}-T_U-T_D) C}{\beta}\!-\!k_3z$, $k_2$ to $k_5$ are constants related to LLM parameters. The feasible solution to P2 with the maximum $z$ is exactly the optimal solution to P1. We propose an efficient algorithm to search for the optimal solution to P1, presented as Algorithm 1. Observing that (2c), (2d) and (2e) increase with $n_i$, we prioritize requests with lower $n_i$ and higher $\tilde{\tau_{i}}$ to minimize constraint violation. Starting with the largest $z$ (Line 2), we rank requests by $\tilde{\tau_{i}}$ and assess P2's feasibility among the top $d$ requests, reorded by $n_i$ (Line 3-5). The approach expedites the solution through sequential prioritization.
\begin{figure}[!t]
	\centering
	\includegraphics[width=3.6in]{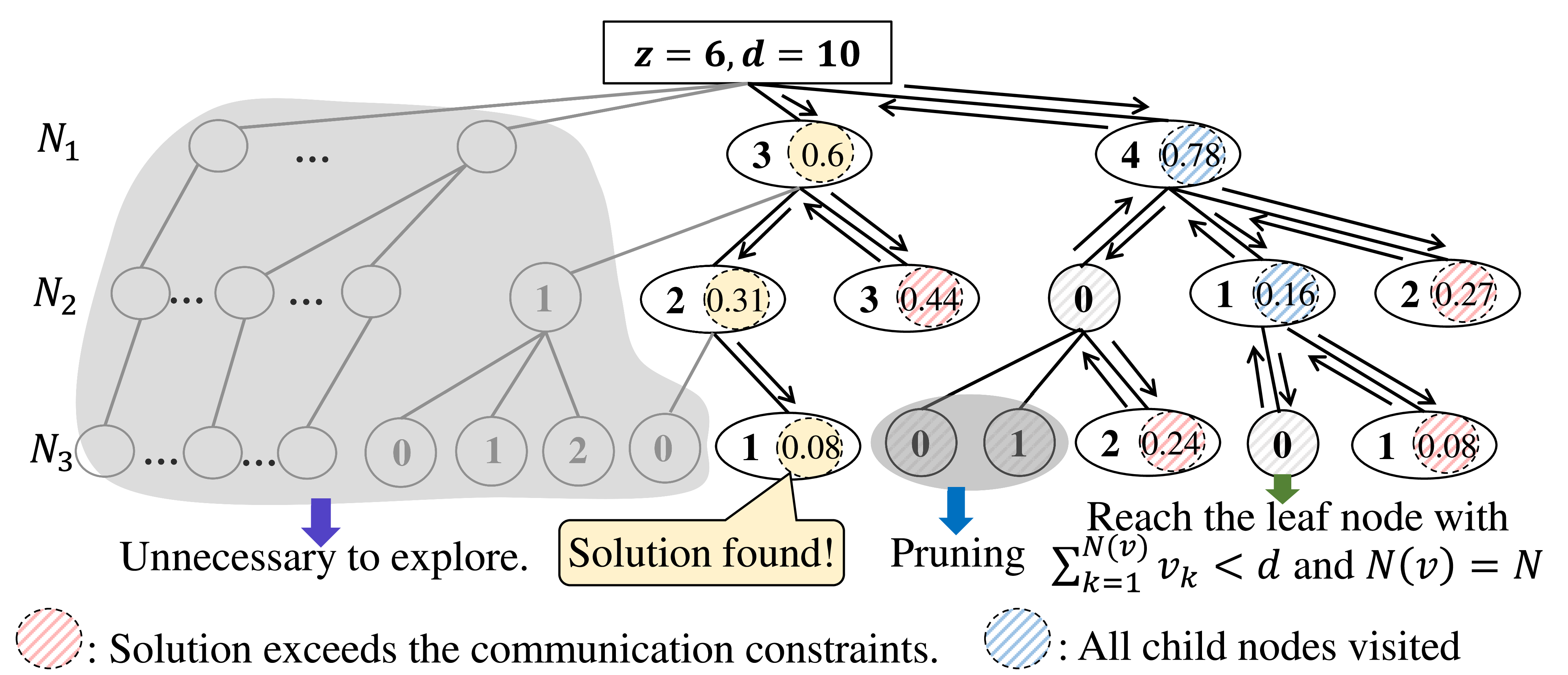}
	\vspace{-5pt}
	\caption{An example of Algorithm 1. $z=6$, $d=10$, $N=3$, $\left|\mathcal{F}_{N_1}\right|=\left|\mathcal{F}_{N_2}\right|=4, \left|\mathcal{F}_{N_3}\right|=2$. All paths meet the memory and latency constraints. Inside each dotted circle, the number represents cumulative uplink bandwidth of requests in $\mathcal{S}_k$. }
	\vspace{-10pt}
\end{figure}
\subsection{Tree Construction}

Given the $z$ and $d$ fixed, the top $d$ candidate requests are denoted as $\mathcal{F}_d$. We categorize $\mathcal{F}_d$ based on output length: $N_1$ represents the shortest. $\mathcal{F}_d =\mathcal{F}_{N_1} \cup \mathcal{F}_{N_2} \cup ... \cup \mathcal{F}_{N} $, where $\mathcal{F}_{N_k} = \left\{i | i \in \mathcal{F}_d, n_i = N_k\right\}, k \in \left\{1,2,...,N \right\}$. Let $\mathcal{S}'$ define the set of selected $z$ requests, $\mathcal{S}'=\mathcal{S}'_1 \cup \mathcal{S}'_2 \cup ... \cup \mathcal{S}'_N$, where $\mathcal{S}'_k = \left\{ i | i \in \mathcal{F}_{N_k}\right\}, k \in \left\{1,2,...,N \right\}$. Each $\mathcal{S}'_k$ is a subset of $\mathcal{F}_{N_k}$. With $\mathcal{F}_d$ established, we aim to identify a potential solution $\mathcal{S}'$ efficiently by constructing a search tree. As shown in Fig. 4, the methodology for this construction is as below.

\textit{(1) Root, parent, and child nodes:} The root node, $v_0$, has child nodes representing possible choices for $\mathcal{S}_1$. While $\mathcal{S}_1$ can be any subset of $\mathcal{F}_{N_1}$, priority is given to requests with lower uplink bandwidth. Each $\mathcal{S}_1$ acts as a parent for $\mathcal{S}_2$, spawning $\left| \mathcal{F}_{N_2}\right|+1$ child nodes, building the search tree layer by layer.

\textit{(2) Path:} Each node $v_{N(v)}$ at depth $N(v)$ uniquely indicates the size of $\mathcal{S}_{N(v)}$. The path from the root to $v_{N(v)}$ can be traced by the vector $\bm{p}=\left[v_0,v_1,...,v_{N(v)}\right]$, representing a partial solution $\mathcal{S}$ where $\left| S_{N(k)} \right| = v_k$ for $k = 1,2,..., N(v)$.

\textit{(3) Leaf node:} A node $v_{N(v)}$ is a leaf if $\sum_{k=1}^{N(v)} v_k = z$, representing a solution, or if $\sum_{k=1}^{N(v)} v_k < z$ and $N(v)=N$, indicating maximum depth without solution. Otherwise, $v_{N(v)}$ has undiscovered child nodes.

\subsection{Tree Searching and Tree Pruning}

Consider a search over the tree constructed for given $z$ and $\tau_{min}$. When dealing with large values of $z$ and $d$, it is impractical to store the entire tree in memory and check the feasibility of all paths. We addresses this by optimizing node exploration order and reducing unnecessary search complexity. 

\textit{(1) Tree-searching:} When exploring all child nodes of any node, we prioritize nodes with the highest index, favoring requests with smaller $n_i$ values. This ensures constraints (2c)-(2e) are met efficiently. Additionally, we emphasize depth over breadth, quickly accessing leaf nodes. Upon exploring a node, we proceed to its largest unvisited child, deepening the search. If a node is terminal or fully explored, we backtrack.

\textit{(2) Tree-pruning:} For any node $v_{N(v)}$, if $\sum_{k=N(v)}^N \left | \mathcal{F}_{N_k} \right | <z - \sum_{k=1}^{N(v)} v_k$, we skip $v_{N(v)}$, its child nodes, its sibling nodes with smaller indices and their child nodes.

\textit{\textbf{Complexity.}} Sorting nodes by uplink bandwidth has a complexity of $\!O(I^2)\!$. Each path's complexity is $\!O(I^2\max \left\{{I,N}\right\})\!$. With the search tree having $\!O((\frac{I}{N})^N)\!$ nodes, the \textit{DFS} complexity is $\!O(I^2\max\left\{{I,N}\right\}(\frac{I}{N})^N)\!$. The overall complexity of Algorithm 1 is $\!O(I^4\max\left\{{I,N}\right\}(\frac{I}{N})^N)\!$. Since $N$ is a constant much smaller than $I$, the complexity is polynomial in the number of $I$, even without considering the complexity ruduction by pruning.

\vspace{-5pt}
\section{Simulations}
\vspace{-3pt}
\begin{table}
	\scriptsize
	\centering
	\renewcommand\arraystretch{1.1}
	\caption{LLM settings in the simulation \cite{zhao2023survey}}
	\vspace{-5pt}
	\begin{tabular}{c | c |c|c|c}
		\hline
		\textbf{Model} & \makecell{\textbf{Layer}\\ \textbf{ number}} & \textbf{Dimension} & \makecell{\textbf{Head}\\ \textbf{number}} & \makecell{\textbf{Head} \\ \textbf{dimension} }\\
		\hline
		BLOOM-3B & 30 & 2560 & 32 & 80\\
		\hline
		BLOOM-7.1B & 30 & 4096 & 32 & 128\\
		\hline
		OPT-13B & 40 & 5120 & 40 & 128\\
		\hline
	\end{tabular}
\end{table}
\begin{table}
	\scriptsize
	\centering
	\renewcommand\arraystretch{1.1}
	\caption{PPL degration under different quantization methods \cite{yao2023comprehensive}}
	\vspace{-5pt}
	\begin{tabular}{c | c |c|c|c}
		\hline
		\textbf{Precision} &  \textbf{Quantization method}  & BLOOM-3B & BLOOM-7.1B & OPT-13B \\
		\hline
		\multirow{2}*{W4A16} & GPTQ & 0.75 & 0.54 & 0.2 \\
		\cline{2-5}
		~ & ZQ-Local & 0.92 & 0.59 & 0.42 \\
		\hline
	\end{tabular}
\vspace{-10pt}
\end{table}

The arrival of user requests are modeled by a Poisson process with rates between 5 to 250 requests per second. The input and output length randomly vary among $\left\{128, 256, 512\right\}$ tokens. We use Byte-Pair Encoding (BPE) tokenization, with each token as a 2-byte index. Latency requirements are uniformly distributed among $\left[0.5,2\right]$ seconds, and the required accuracy range from $\left[0,1\right]$ PPL degradation. In simulations, our edge server uses 20 NVIDIA JETSON TX2 GPUs, each with 1.33TFLOPs and 32GB memory. Bandwidth is 20MHz. Transmit powers are 20dBm (user-to-EN) and 43dBm (EN-to-user) \cite{9576659}. $N_0 =$ -174dBm/Hz. The channel gain follows Rayleigh fading at $10^{-3}$ pass loss \cite{10038543}. The duration of an epoch is set as 2 seconds, with $T_U=T_D=$250ms. Two benchmarking schemes of batched inference are adopted. (1) Static batching (StB): The edge node has a set batch size based on epoch duration and LLM parameters to avoid GPU overflow. (2) No batching (NoB): Each GPU accepts a request once idle. Three open-source LLMs are adopted in this simulation, as detailed in Table 1. The FFN's dimension is four times the model's. Default quantization is 8-bit weight, 16-bit activation (W8A16). Other methods are in Table 2.

Fig. 5 contrasts the batching schemes between BLOOM-3B and BLOOM-7.1B inference. In Fig.5(a), throughput initially increases with the arrival rate but later stabilizes due to edge node constraints. The DFTSP scheme surpasses both StB and NoB. This is because DFTSP dynamically adjusts batch sizes, while StB maintains a consistent batch to prevent overflow and NoB handles one request per GPU. Overall, BLOOM-7.1B has a reduced throughput compared to BLOOM-3B across all schemes, attributed to its larger parameter count.
\begin{figure}[!t]
	\centering
	\includegraphics[width=3.4in]{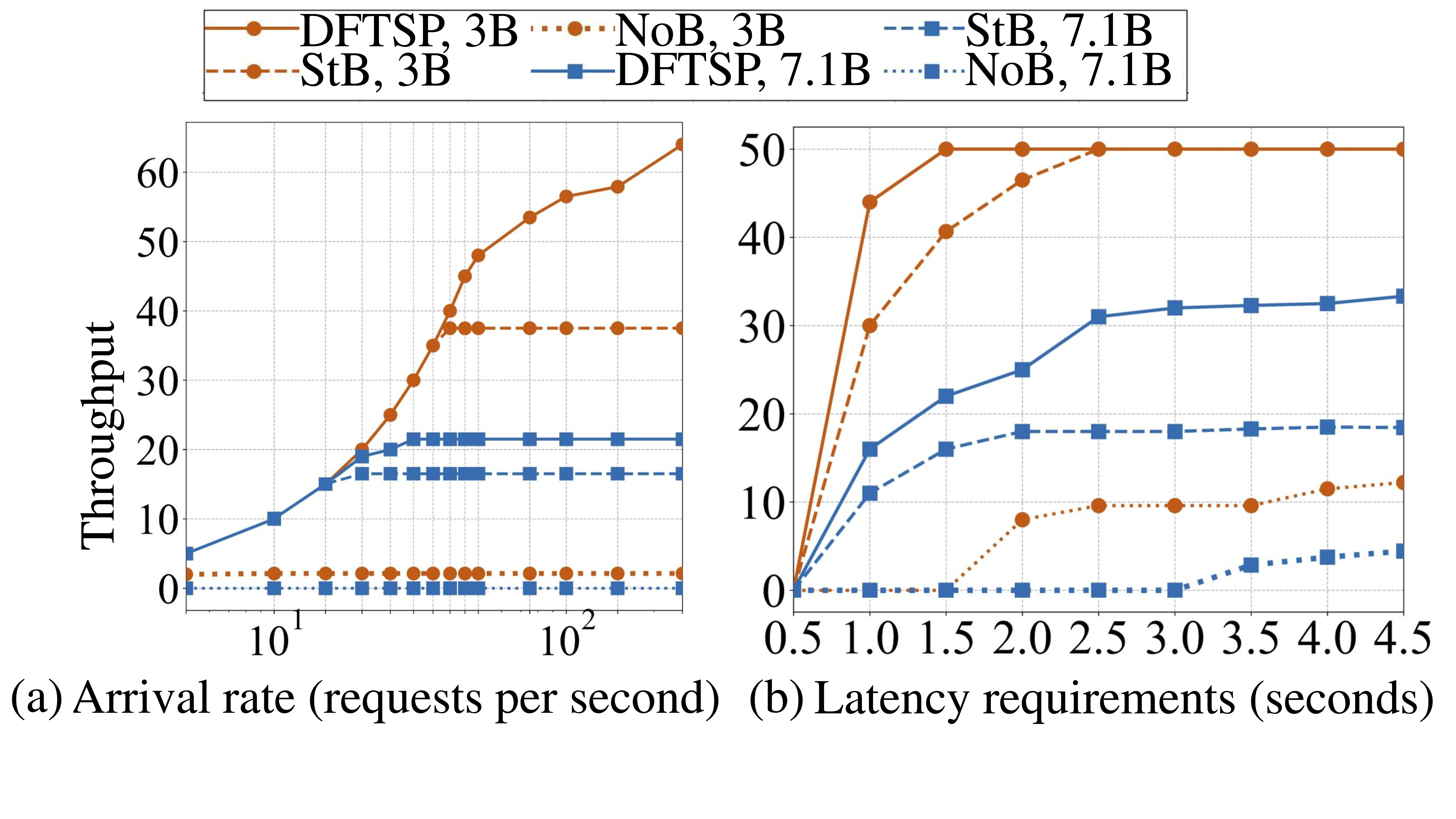}
	\vspace{-15pt}
	\caption{Throughput under different batching schemes. }
	\vspace{-6pt}
\end{figure}
\begin{figure}[!t]
	\centering
	\includegraphics[width=3.4in]{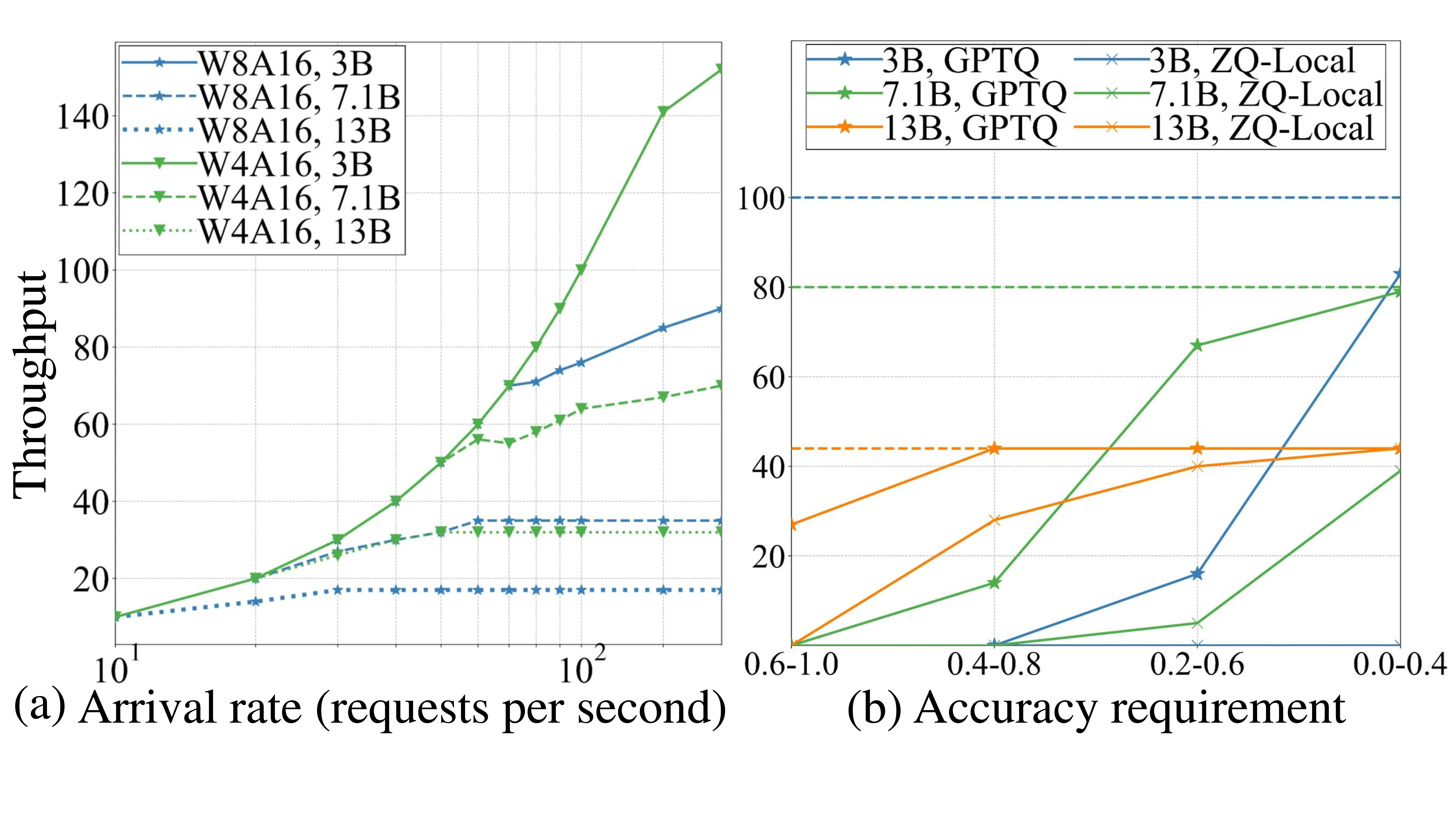}
	\vspace{-15pt}
	\caption{Throughput under different quantization methods. }
	\vspace{-15pt}
\end{figure}
In Fig. 5(b), throughput varies based on user latency requirements. As these requirements become more lenient, more requests meet their deadlines. BLOOM-3B consistently surpasses BLOOM-7.1B due to its optimized dimensions and reduced matrix operations. NoB, without the advantage of parallel processing, struggles to consistently meet tight deadlines for larger BLOOM-7.1B inferences.

Fig. 6 delves into the throughput effects of various quantization methods. In Fig. 6(a), overlooking user accuracy requirements, we observe that larger models using the same quantization precision handle fewer requests per epoch, constrained by computational resources. However, diminishing quantization precision minimizes memory usage, boosting throughput. Still, there are trade-offs: Fig. 6(b) highlights that broader quantization can expand the PPL differential, thus affecting accuracy and subsequent throughput. Dotted lines denote W8A16 quantization throughput. Intriguingly, even at the same precision, GPTQ and ZQ-Local quantizations vary due to their distinct tensor operations. This suggests the importance of choosing the right quantization method tailored to specific LLMs. As accuracy constraints are relaxed, more inferences are processed.

Lastly, we contrast the computational complexity of the DFTSP algorithm, optimized for tree-search, with a benchmark that lacks tree-pruning. Table 3 underscores that brute-force search complexity surges exponentially with higher arrival rates. The merits of DFTSP become even more apparent with rising rates, achieving a notable 97.92\% complexity reduction at an arrival rate of 200.

\vspace{-4pt}
\section{Conclusion}
\vspace{-2pt}

In this paper, we delved into the potential of edge intelligence in the context of LLM inference. We investigated the optimization of edge intelligence tailored for LLM inference in wireless networks, focusing on the joint optimization of batching scheduling and resource allocation to maximize throughput. To balance the trade-off between quantization and LLM service accuracy, we introduced the PPL differential metric. We developed the DFTSP algorithm, which efficiently achieves the optimal solution in polynomial time using tree-search and tree-pruning techniques. Simulation results underscored DFTSP's superior throughput enhancement compared to other batching approaches. These findings also shed light on selecting appropriate quantization schemes for specific LLMs in resource-constrained wireless edge nodes. Notably, DFTSP significantly outperforms the brute-force tree-search algorithm on time complexity.

\begin{table}
	\scriptsize
	\centering
	\renewcommand\arraystretch{1.1}
	\caption{Algorithm Time Reduction with Tree-Pruning}
	\vspace{-5pt}
	\begin{tabular}{c | c |c|c|c}
		\hline
		\makecell{\textbf{Arrival rate} \\ \textbf{(requests per second)}}&10&50&100&200 \\
		\hline
		\textbf{Complexity Reduction}& 45.52\% & 71.18\% & 79.07\% & 97.92\% \\
		\hline
	\end{tabular}
	\vspace{-10pt}
\end{table}

\vspace{-4pt}
\bibliographystyle{IEEEtran}
\footnotesize
\bibliography{mybib}

\end{document}